\documentclass{article}

\usepackage[final]{nips_2016}
\usepackage{natbib}
\setcitestyle{numbers}
\usepackage[utf8]{inputenc} 
\usepackage[T1]{fontenc}    
\usepackage{hyperref}       
\usepackage{url}            
\usepackage{booktabs}       
\usepackage{amsfonts}       
\usepackage{nicefrac}       
\usepackage{microtype}      
\usepackage{graphicx}
\usepackage{fancyhdr}
\usepackage{amsmath}
\usepackage{todonotes}

\fancypagestyle{equalc}{\fancyhf{}\fancyfoot[R]{* indicates equal contribution, full list of team member contributions available at \url{github.com/PsiPhiTheta/catch-me-if-you-can/blob/master/CONTRIBUTING.md}}}

\title{Catch Me If You Can}

\author{
    C. J. Li* \\
    Department of Computer Science\\
    University of Toronto\\
    Toronto, ON M5S 3H7 \\
    \texttt{casey@cs.toronto.edu} \\
    \And
    A. Viscardi* \\
    Department of Computer Science\\
    University of Toronto\\
    Toronto, ON M5S 3H7 \\
    \texttt{avis@cs.toronto.edu} \\
    \And
    T. Hollis* \\
    Department of Computer Science\\
    University of Toronto\\
    Toronto, ON M5S 3H7 \\
    \texttt{thollis@cs.toronto.edu} \\
}

\begin{document}

\maketitle

\begin{abstract}

As advances in signature recognition have reached a new plateau of performance at around 2\% error rate, it is interesting to investigate alternative approaches. The approach detailed in this paper looks at using Variational Auto-Encoders (VAEs) to learn a latent space representation of genuine signatures. This is then used to pass unlabelled signatures such that only the genuine ones will successfully be reconstructed by the VAE. This latent space representation and the reconstruction loss is subsequently used by random forest and kNN classifiers for prediction. Subsequently, VAE disentanglement and the possibility of posterior collapse are ascertained and analysed. The final results suggest that while this method performs less well than existing alternatives, further work may allow this to be used as part of an ensemble for future models. 
  
\end{abstract}

\thispagestyle{equalc}

\section{Introduction}

Signatures are one of the oldest accepted biometric markers of identity in use today. Despite their prevalence in modern legal and financial systems, signatures are poorly authenticated by common workflows and are prone to forgery. Check fraud costs banks around the world over \$900M a year, with 22\% of these frauds being assigned to signature forgery. With over 27.5 billion checks circulating through the US only in any given year, the automation of signature verification has become a priority for governments around the world.

There are three types of forgery: random forgery, simple forgery and skilled forgery. Random forgery is when the forger does not have access to the authentic signature or the name of the person. Simple forgery is when the forger knows the name but not the signature of the person to imitate. Skilled forgery is when the forger knows the authentic signature and can use it to train his forgery with. This latter type of forgery is the one that is particularly difficult to discern and which has been the subject of vast academic research for over a century.

Two major signature verification competitions in the last two decades (ICDAR SigComp11 and SVC2004) have open sourced significant volumes of signature data. This has resulted in the development of algorithms which achieve at best around 2\% error rate. This error rate, while drastically better than human alternatives, still leaves much work to be done. The aim of this paper is to explore alternative approaches that have not yet been investigated to shed more light on strategies that may help defeat this 2\% `gold standard' of signature verification.

The main idea of this paper is to use variational auto-encoders (VAEs) to try to undertake one-class classification on signature data. This may nonetheless need a lot of data in order to produce a strong generative model so we decided to investigate this. We also investigated some few-shot-learning approaches to help tackle this problem with a more realistic limited training set. One hypothesis, inspired by MAML \cite{finn2017model}, is that we may be able to train a set of latent parameters on a large set of general human signatures which will generalize quickly to fit a specific person's signature.

Signature data can be either offline data (2D image of the resultant signature) or online (dataset of live features such as velocity of tip, pressure, time...). While there is an abundance of signature data available, the main dataset that we will use for offline data is the ICDAR competition dataset \cite{liwicki2011signature}. Although out of scope of this report, an important dataset identified for online data is the SVC2004 competition dataset \cite{yeung2004svc2004}. 

The VAE approach coupled with a classifier, here k-NNs and random forests, constitutes the main contribution of this paper. This is an example of anomaly detection by one-class classification. "Anomalous" inputs, in our case forged signatures, will map to a space that falls outside of some decision boundary \cite{ruff2018deep}. In addition to this, VAE disentanglement and the possibility of posterior collapse are also ascertained and analysed. 

\section{Related Work}

While the origins of Auto-Encoder (AE) neural networks remain a disputed topic, auto-encoders seem to first be proposed as a method for unsupervised pre-training in \cite{ballard1987modular}. As the introduction in \cite{ballard1987modular} explains, it's somewhat difficult to attribute origins to all the ideas within neural network research as the literature is spread across various fields and terminology has substantially evolved over time. Since then, various classes of modifications of the vanilla multilayer AEs have been developed to tackle a wide array of problem domains. Indeed, the AE taxonomy is so vast that it out of scope of this paper, but a brief overview of the important classes in handwriting-related computer vision remains important to outlining this work’s background literature. 

Firstly, the AE concept of a bottleneck layer has been applied to convolutional neural networks (CNNs) to form the class of convolutional auto-encoders (CAEs) \cite{masci2011stacked}.  In \cite{masci2011stacked}, Masci et. al. apply a stack of CAEs to the MNIST \cite{lecun2010mnist} and CIFAR-10 \cite{krizhevsky2014cifar} benchmark datasets, achieving improved results. 

Another notable class of AE modifications is regularised autoencoders. This class of AE modifications constrains the neural network into a bottleneck by regularising them rather than by using a size limit. Examples of this class of AEs includes the Sparse AE \cite{makhzani2013k} and the Denoising AE \cite{vincent2010stacked}. The Sparse AE leverages a sparsity penalty in its loss function to constrain the hidden unit reconstruction from the latent space representation. On the other hand, the Denoising AE inserts noise to the input image to force the autoencoder bottleneck to learn a latent space representation of important features rather than overfitted ones. Both of these architectures have been shown to yield significant improvements on handwriting-based supervised learning tasks \cite{makhzani2013k} \cite{vincent2010stacked}.

However, the main class of AEs concerned in this paper is variational Auto-Encoders (VAEs), first proposed by Kingma and Welling in the seminal paper ``Auto-Encoding Variational Bayes'' \cite{kingma2013auto}. In \cite{kingma2013auto}, Kingma outlines the modification that can be done to regular AE architectures to allow them to also become generative models. This modification consists in replacing the bottleneck vector by two vectors, one representing the mean of the distribution while the other represents the standard deviation, allowing the creation of a latent distribution that can be sampled from. Since then, this architecture has been successfully leveraged against various anomaly-detection problems, often yielding very encouraging results. Further sub-modifications within this class of AE have also yielded interesting results such as the Disentangled VAEs proposed by Google DeepMind in ICLR 2017 \cite{higgins2017beta}. This architecture tries to make sure different neurons in the latent distribution are uncorrelated, thereby each trying to learn something different from the input data. 

As for existing literature on signature verification, this is a plentiful research space of academia. Most of the work that has been done can be split into two categories: pre-2004 and post-2004. 2004 marks an important year in this field as it is when the first significant global competition was released to try to tackle this problem in an automated manner. 

A really informative review of historical methods pre-2004 can be found in the Pattern Recognition paper by Plamondon and Lorette \cite{plamondon1989automatic}. This paper outlines the general fields at the time which were divided amongst text and signature recognition, further subdivided by recognition, identification and verification techniques (usually using distance metrics such as euclidean distance, squared distance or Mahalanobis distance). One of the most well known publications of this era is the NIPS paper from LeCun on Siamese time delay Neural Networks for signature verification \cite{bromley1994signature}.

As for later methods, the most performant modern approach to online data (using the SVC2004 dataset) was by Fayyaz et. al. in \cite{fayyaz2015online}. In \cite{fayyaz2015online}, one-class classifiers are leveraged by learning features from users’ signatures using autoencoders. This outperforms the original winners of the SVC2004 competition \cite{kholmatov2005identity} who used a linear classifier with PCA to achieve at 2.8\% error rate. Conversely for offline data, we have a benchmark using autoencoders that reaches a classification AUC of around 0.9 (which varies with the skill of the forger) \cite{souza2008combining}.

On the other hand, meta learning is another highly active area of research that aims to improve the performance of existing learning algorithms by learning good parameters for those learners. Model-Agnostic Meta Learning (MAML), as proposed by Finn, Abbeel and Levine in \cite{finn2017model}, is one such algorithm that is able to run on top of any model trained using gradient descent on any problem from classification to regression and even reinforcement learning. In our case, this few-shot-learning algorithm is particularly desirable for signature verification as we may have a very limited opportunity to detect forgeries and would like to do so using the least amount of training iterations. Another possibility for meta-learning in the signature verification problem domain is to learn a one-shot learner. As presented in \cite{bertinetto2016learning}, learning feed-forward one-shot learners can be done by using a secondary deep network called a learnet which predicts the pupil network’s parameters from a single exemplar. While zero-shot learning has been gaining notable traction since the NeurIPS 2013 paper on the subject \cite{socher2013zero}, this does seem to be something of promising potential in the field of signature verification. This is because it would assume that someone’s signature can be modelled using only examples of other people’s signatures or other people’s forgeries, which seems like a bit of a stretch for now. Existing work in meta-learning for signature verification is quite limited although there is one paper that uses an HMM algorithm for one-shot learning \cite{muramatsu2003hmm} of online signature verification. This work, undertaken in 2003, does seem somewhat outdated so it would be interesting to see if modern approaches to signature verification can benefit from aforementioned few-shot learning methods.

The principal contribution in this paper is the detailed study and leveraging of VAEs for signature verification. While the implications of this for using VAEs with MAML was briefly investigated, it was deemed more relevant to study posterior collapse as detailed hereafter. 

\section{Model Architecture}

The main model architecture developed in this paper was a modified VAE that parses augmented signature images, based on an MNIST VAE \cite{Keras2018VAE}. A diagram of this modified VAE architecture is shown in Figure \ref{system-diagram} below.

\begin{figure}[!h]
    \begin{center}
    	\includegraphics[width=0.95\linewidth]{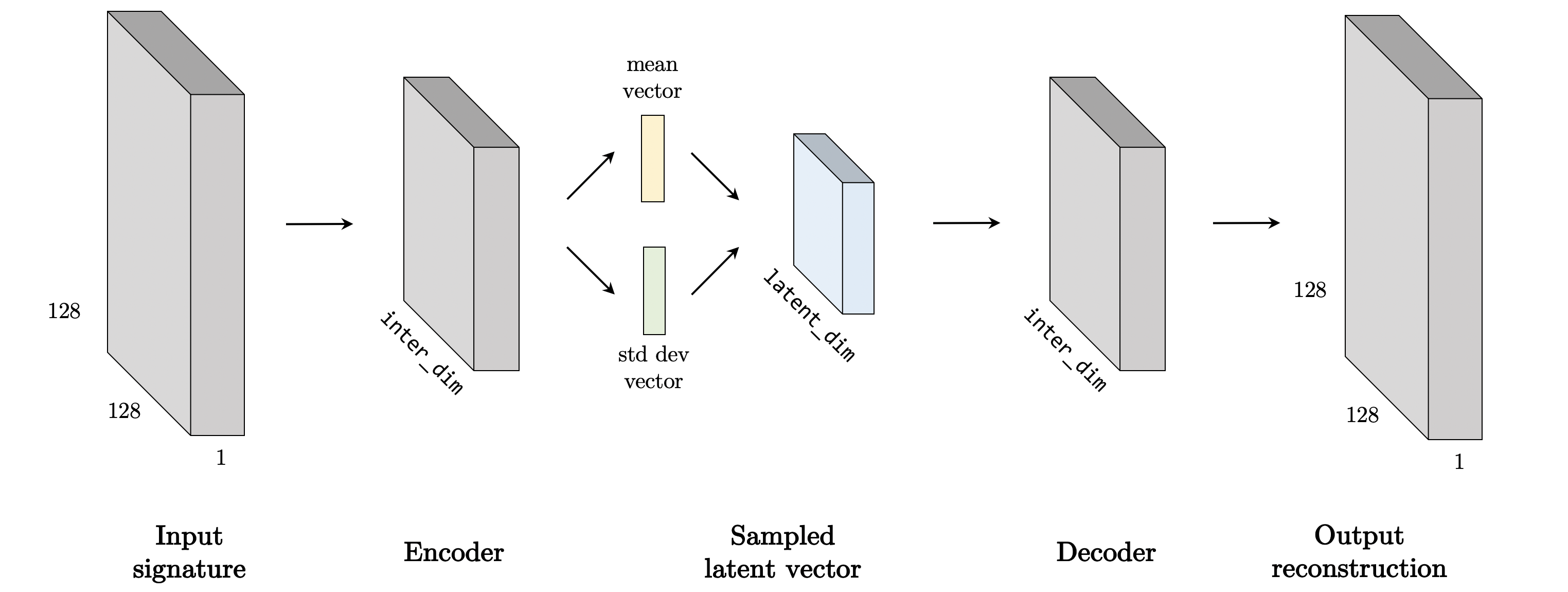}
    	\caption{System diagram of the VAE architecture used}
        \label{system-diagram} 
    \end{center}
\end{figure}

As seen in Figure \ref{system-diagram}, VAEs are a special type of AE that can also be used as a generative model, by modeling the latent representation as a distribution that can be sampled, rather than a deterministic mapping. The main difference between VAEs and regular AEs is that the bottleneck layer is replaced by the sampled latent vector which statistically approximates a distribution using a mean and standard deviation vector. The decoding of outputs from this distribution is simultaneously optimized to reconstruct the input.   

Mathematically, let us define a prior $p(z)$ over the latent code $z$ and a likelihood on the data, $x$, expressed as $p(x|z)$ such that it is conditioned on the latent $z$ which we are seeking to learn. The model thus defines a joint probability $p(x,z)=p(x|z)p(z)$. The objective is to learn a code value $z$ that encodes a useful representation of $x$ for some purpose. In our case, the latent space objective of inferring the code value $z$ given an observed datapoint $x$ can be formulated via Bayes Theorem as \cite{jaan}: 
\begin{align}
    p(z | x)=\frac{p(x | z) p(z)}{p(x)}    
\end{align}
Marginalizing over $z$ to obtain the denominator is usually intractable as it would require integration over all possible configurations of $z$, which becomes exponentially large with increased dimensionality of $z$. Thus the true $p(z|x)$ is usually approximated using a distribution $q_{\phi}(z | x)$ parameterized by $\phi$. In order to ensure that the variational posterior $q(z|x)$ approximates the true posterior $p(z|x)$, we can use KL divergence. As the goal is to find variational parameters $\phi$ that minimize the KL divergence between the true and variational posteriors, we have:
\begin{align}
    q_{\phi}^{*}(z | x)=\arg \min _{\phi} \mathbb{K} \mathbb{L}\left(q_{\phi}(z | x) \| p(z | x)\right)
\end{align}
However, this remains intractable as the evidence $p(x)$ remains in the divergence. To address this issue we can use the Evidence Lower BOund (ELBO) to allow us to approximate posterior inference. Rewriting the posterior using ELBO yields the following:
\begin{align}
    \log p(x)=ELBO(\phi)+\mathbb{K} \mathbb{L}\left(q_{\phi}(z | x) \| p(z | x)\right)
\end{align}

From (3), we can show by Jensen's inequality that the ELBO for a single datapoint in the VAE is:
\begin{align}
    E L B O_{i}(\phi)=\mathbb{E}_{q_{\phi}(z|x_{i})} [\log p(x_{i}|z)]-\mathbb{K} \mathbb{L}(q_{\phi}(z | x_{i}) \| p(z))
\end{align}

From (4) we can thus show that the reconstruction loss that normal AEs use is now composed of two expressions as follows: 
\begin{align}
    \mathcal{L}(\phi; \mathbf{x}, \mathbf{z}) = \mathbb{E}_{q_{\phi} (\mathbf{z}|\mathbf{x})} [\log p (\mathbf{x}|\mathbf{z})] - \mathbb{K} \mathbb{L}(q_{\phi} (\mathbf{z}|\mathbf{x}) || p(\mathbf{z}))
\end{align}

In (5) we clearly see that the loss has two parts, the reconstruction loss (left) and the KL divergence loss (right). This loss ensures that it minimises the squared error in reconstructing $\mathbf{x}$ from $\mathbf{z}$ while ensuring that the distribution learned is not too far from a normally distributed Gaussian.

The second term is not strictly necessary to achieve the objective, but acts as a regularizer to encourage a meaningful $z$. Otherwise the encoder could simply assign a different region of Euclidean space to each datapoint, to be subsequently reconstructed by a decoder that has simply memorized the inputs, instead of learning truly useful representations in the basis of $z$ \cite{jaan}. The ideal weight for the KL regularization term itself is a highly-researched question as it can affect whether the model suffers from posterior collapse, which is discussed further in Section \ref{section-posterior}. Using appropriately high values of a beta coefficient can also be interpreted as training a VAE to encourage disentangled priors, on the basis of the idea that a latent code which is distributed similarly to a standard Gaussian has de-correlated basis vectors by definition \cite{higgins2017beta}. This would provide a measure of explainability to each latent dimension, and we briefly look at this idea in Section 4.

There is an additional hurdle to overcome in the network architecture, due to our use of gradient descent for optimization: the operations on every node with a trainable parameter have to be differentiable. The sampling operator at the latent layer is stochastic and as such we cannot back-propagate gradients through it. In order to resolve this hurdle and train the network end-to-end using backprop, we make use of the \textit{reparameterization trick}. Mathematically we can consider the sampled latent vector as the sum of the mean with a weighted stochastic standard deviation:
\begin{align}
    z = \mu + \sigma \odot \epsilon
\end{align}

From (6), we can see that by splitting the sampling operation we actually only need to learn the $\mu$ and $\sigma$ during training. As such we can now pass gradients through $\mu$ and $\sigma$ but not through the fixed stochastic $\epsilon$ node. This works because the computation graph is no longer dependent on $\epsilon$ for gradient flow as it has become an isolated node. 

\pagebreak

\section{Experiments}

\begin{figure}[!h]
	\includegraphics[width=0.95\linewidth]{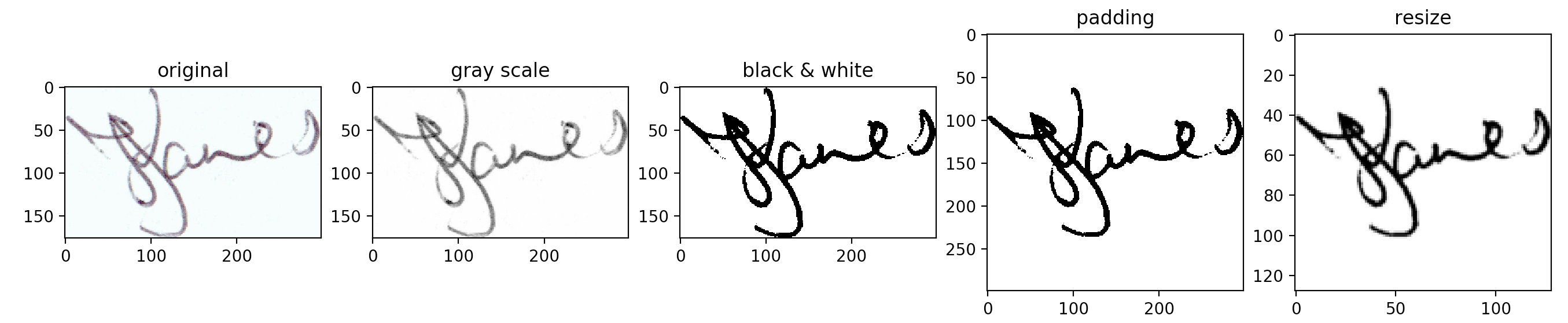}
	\includegraphics[width=0.95\linewidth]{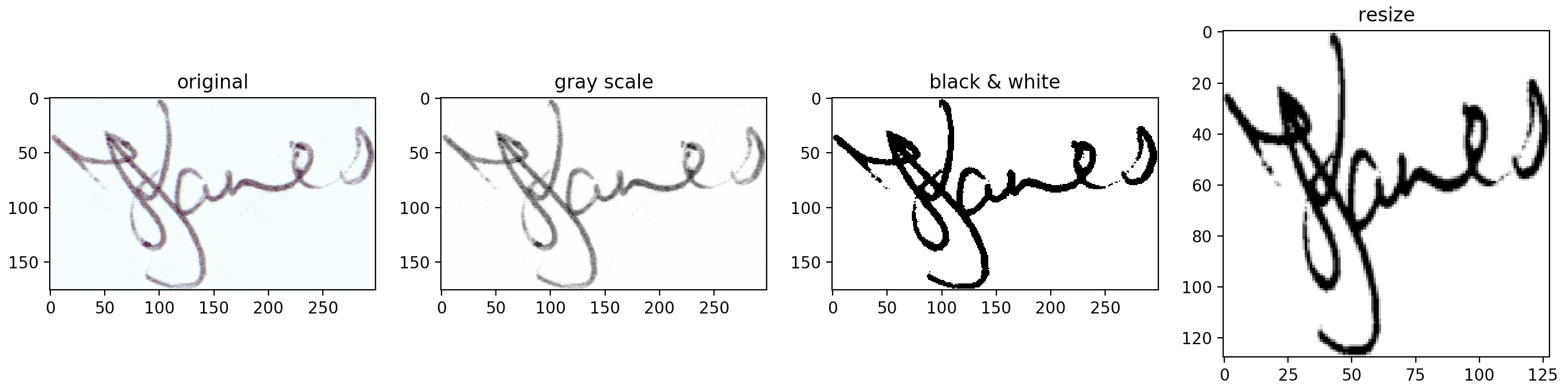}
	\caption{Preprocessing with padding (top) and without padding (bottom)}
	\label{preprocessing} 
\end{figure}

As all of the ICDAR offline signature images are of different sizes and in \texttt{.png} format, they must first and foremost be preprocessed. The preprocessing involves converting the original images to gray scale, black \& white and then optionally padding them before resising to a NumPy array of size $128 \times 128$. The preprocessing pipeline is shown in Figure \ref{preprocessing} as follows. 

\begin{figure}[!h]
    \begin{center}
    	\includegraphics[width=0.73\linewidth]{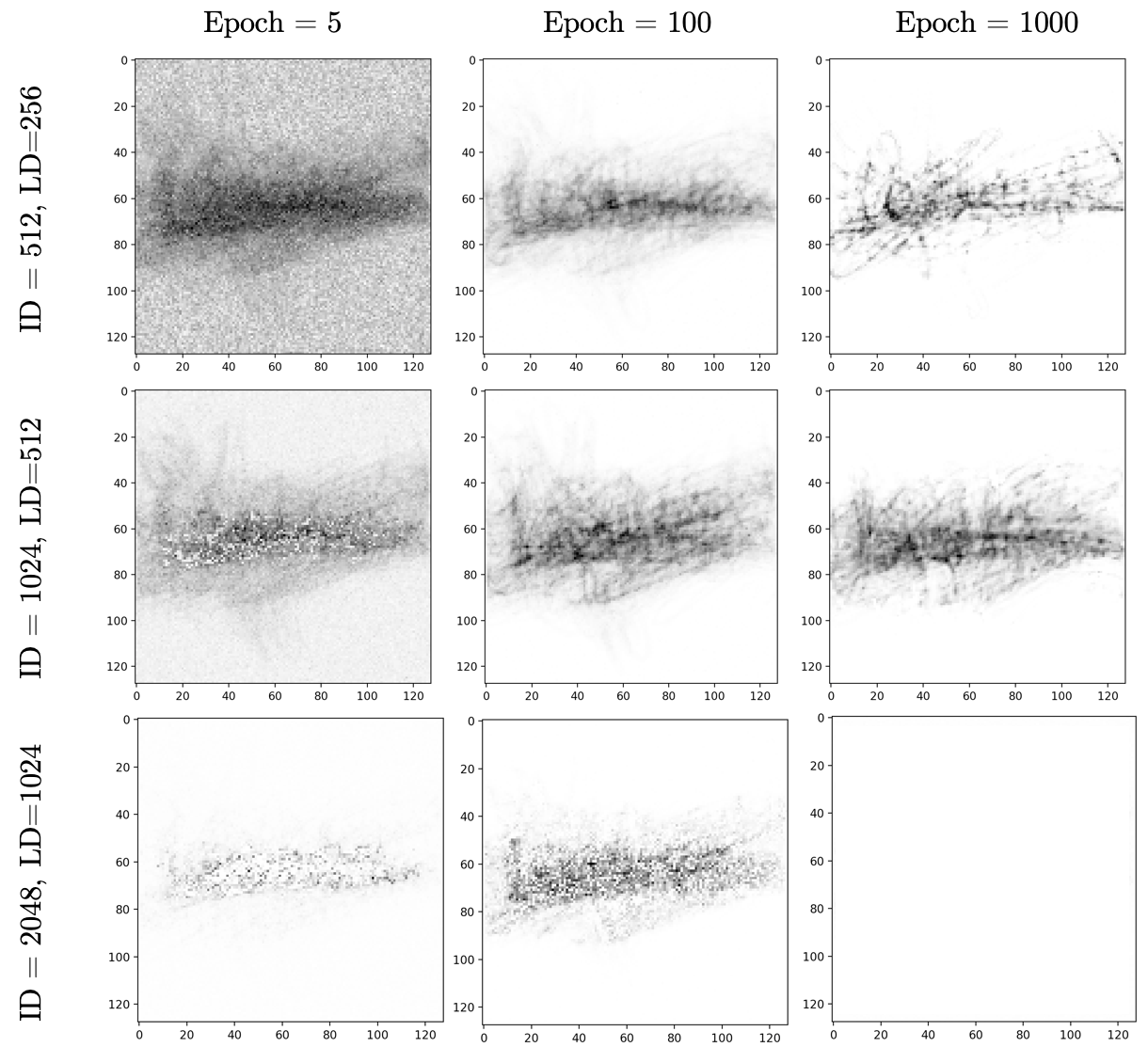} 
        \includegraphics[width=0.24\linewidth]{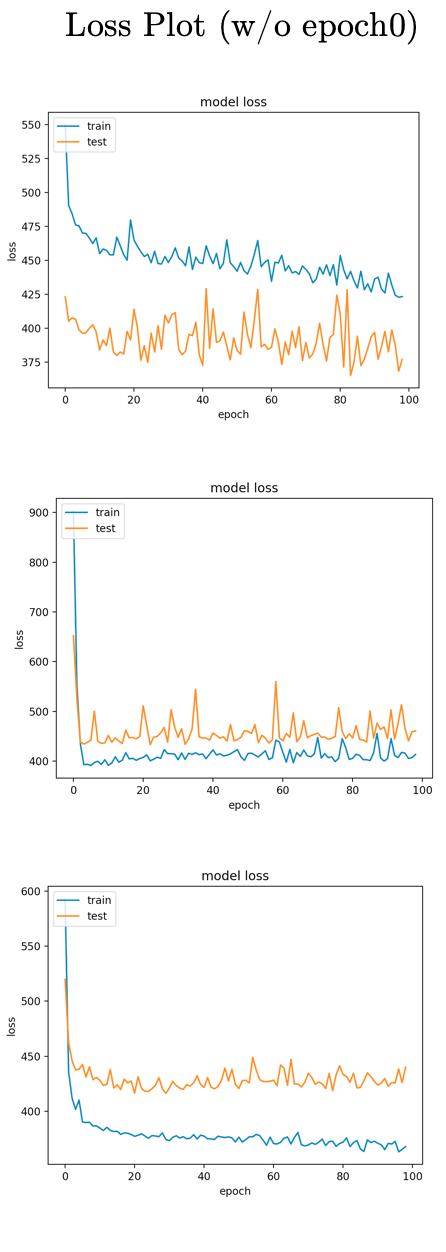}
    	\caption{VAE trained on limited data (all signatures, real and forged)}
    	\label{all-sig-training}
    \end{center}
\end{figure}

Once the ICDAR dataset is preprocessed, a VAE can be trained on the entire ICDAR dataset. The results of this training process by epoch are shown in Figure \ref{all-sig-training}. 

Figure \ref{all-sig-training} shows the loss curves for the first 100 epochs and the latent dimension samples of the trained VAE (3 different sizes) at epoch 5, 100 and 1000. The intermediate dimension, labelled ``ID'', ranges from 512 to 2048 while the latent dimension, labelled ``LD'', ranges from 256 to 1024. Usually, the smaller VAEs ($ID=512$, $LD=256$) achieve better performances and the loss did not significantly decrease beyond epoch 100. Figure \ref{all-sig-training} does however show that the original ICDAR dataset is vulnerable to the small data problem. Indeed, loss curves are very noisy and reconstruction loss is not as good as it could be. More evidence to suggest this can be seen in the low quality examples that were generated by sampling the latent dimension, after 5 and 100 epochs, as seen in Figure \ref{all-sig-training}. This is in line with prior theoretical expectations as VAEs, like many generative models, need a lot of data to produce convincing latent space samples. 

\begin{figure}[!h]
    \begin{center}
    	\includegraphics[width=0.45\linewidth]{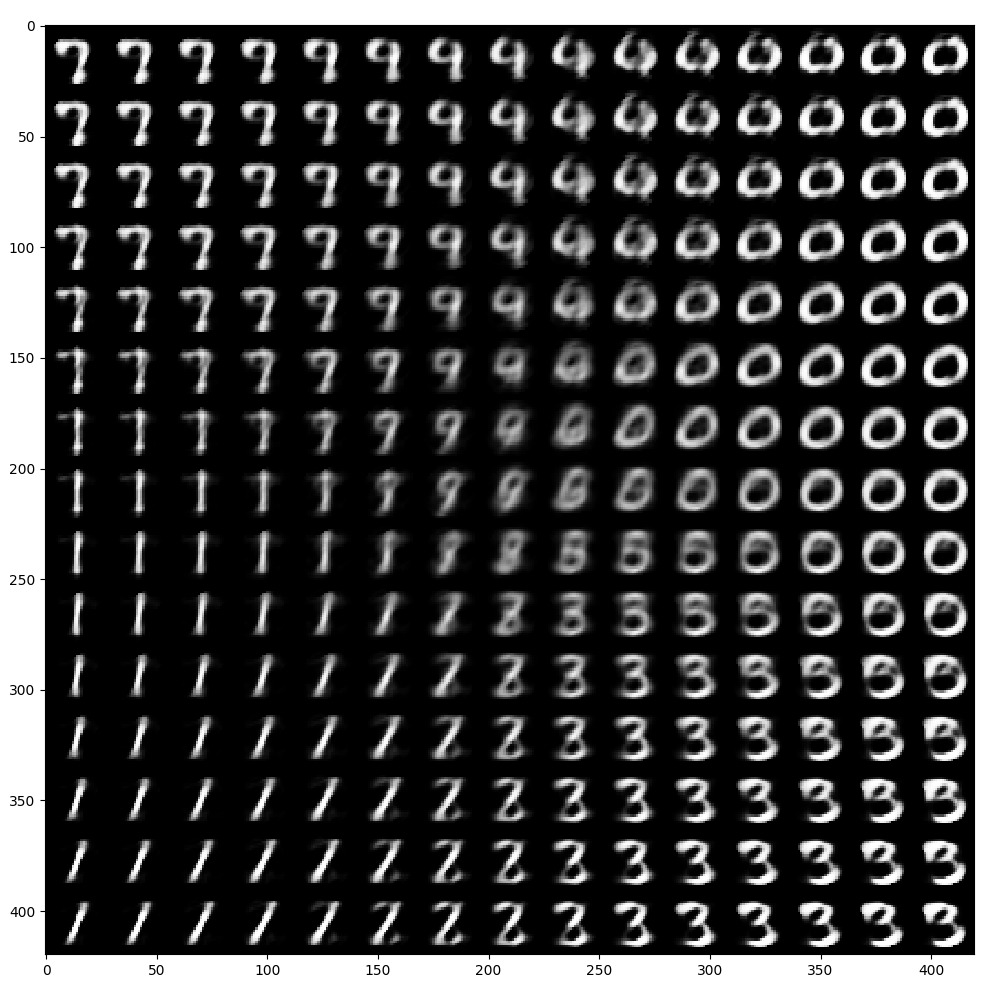}
    	\includegraphics[width=0.45\linewidth]{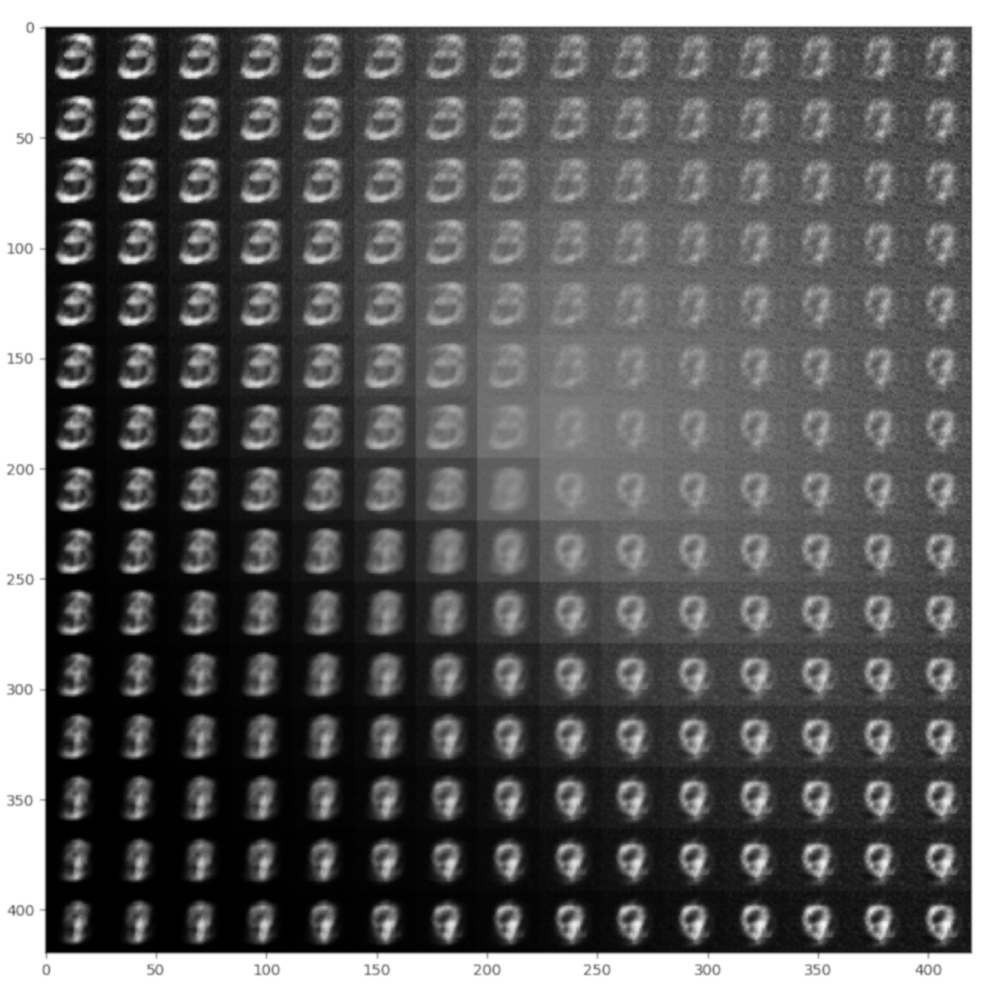}
    	\caption{Digits generated by the VAE trained on the entire MNIST dataset (left) and on a small subset containing 200 images of the MNIST dataset (right)}
    \end{center}
    \label{mnist-1}
\end{figure}

Nevertheless, in order to confirm this claim and rule out the possibility of our sampling process being faulty, we conducted an experiment that consisted in training two similar models on the MNIST dataset \cite{lecun2010mnist} and qualitatively inspecting digits generated by sampling the latent space. Both models featured a 2-dimensional latent space which allowed for a simple and efficient visualisation of the samples from different regions of the latent space. The first model was trained on the entire MNIST dataset while the second was trained on a subset of the MNIST dataset containing 200 images. This subset was chosen as it is similar in size to the ICDAR signature dataset when compared to the input dimensionality. That is, the ratio of input pixels to the number of training examples is approximately equal ($\frac{200}{38\times38} \approx \frac{2000}{128\times128}$) for both the MNIST and ICDAR images allowing for direct comparison. Figure \ref{mnist-1} shows digits generated by sampling the latent space of both networks. The images generated by the network trained on the entire MNIST dataset features sharp edges and recognizable digits, which is in line with the results expected from a generative VAE model. On the other hand, the digits generated by the network trained on a small subset of the MNIST dataset reveal the same type of noise and blurry edges present in the generated signatures of Figure \ref{all-sig-training}. This confirms that our network and the sampling process are not faulty but rather are vulnerable to a small data problem. 

\begin{figure}[!h]
    \begin{center}
        \includegraphics[width=0.48\linewidth]{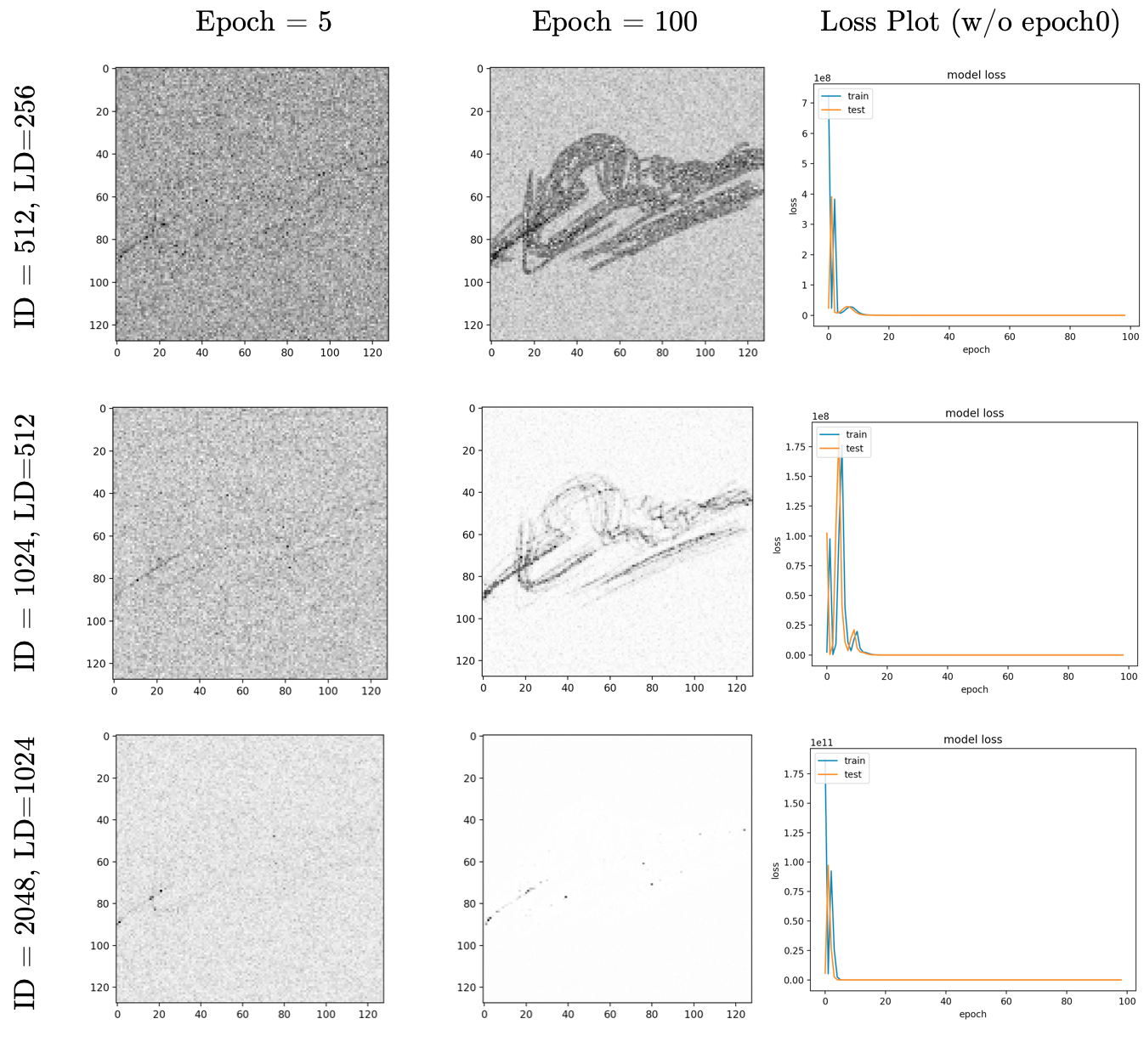} \\
        \includegraphics[width=0.40\linewidth]{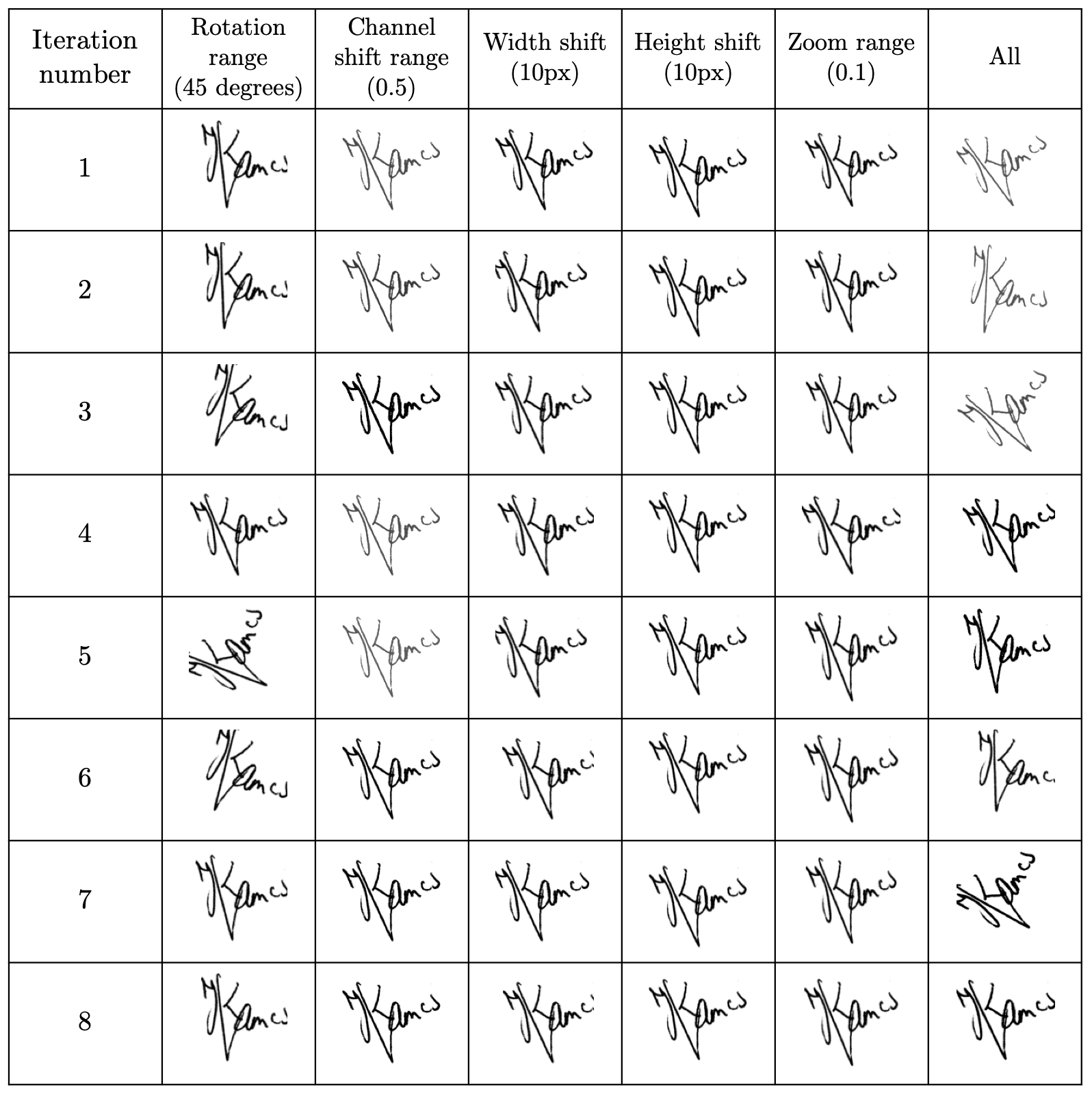} 
        \includegraphics[width=0.48\linewidth]{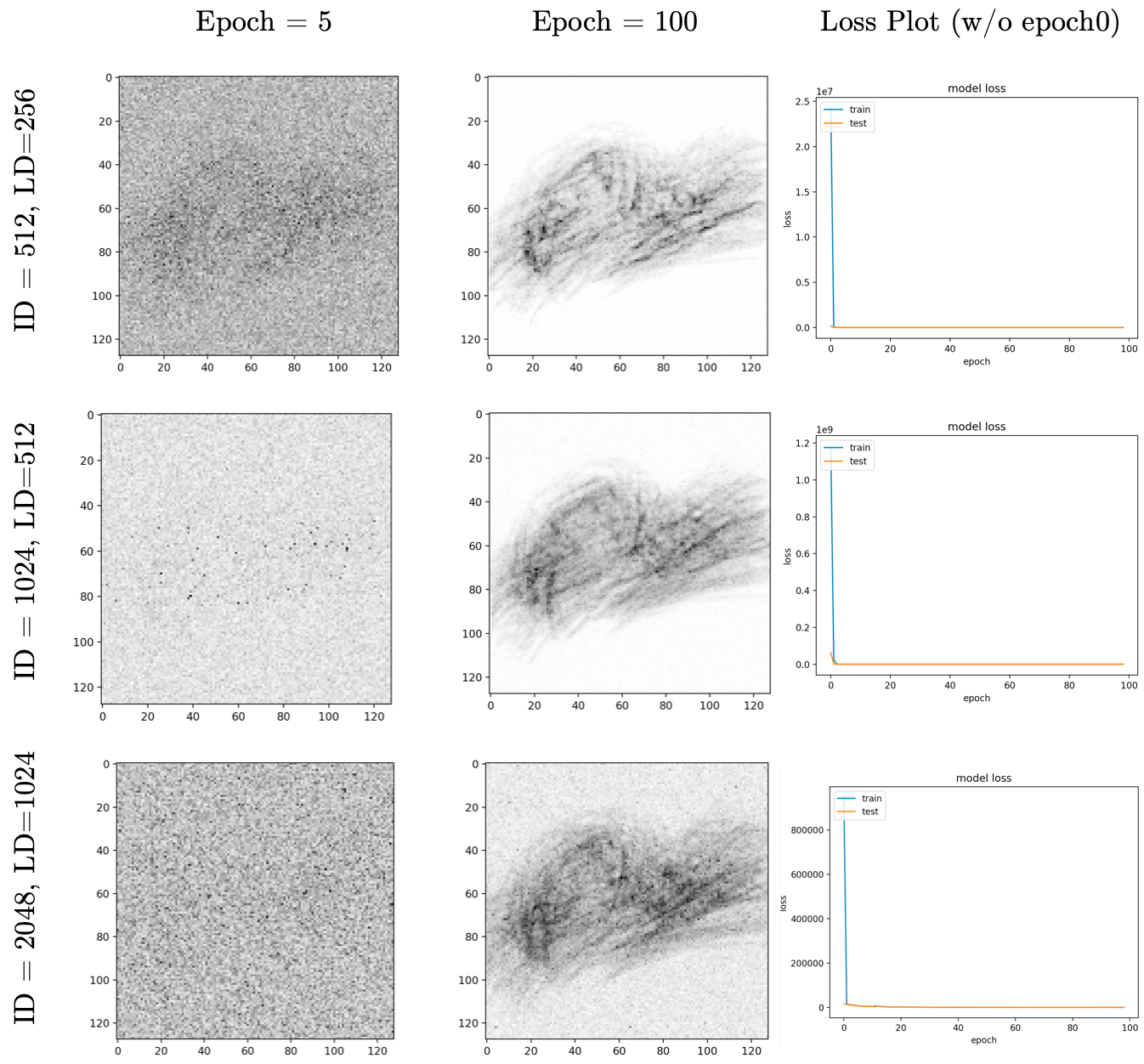}
    	\caption{Impact of data augmentation on VAE training (bottom) w.r.t non-augmented training (top)}
	    \label{aug-training}
    \end{center}
\end{figure}

To attempt to remedy this small data problem, data augmentation was undertaken on the ICDAR offline signature dataset. This allows single signatures to be better generated by the VAE. The results using augmented data compared to non-augmented data are shown hereafter in Figure \ref{aug-training}.

Figure \ref{aug-training} shows that data augmentation was done by combining five different transformations: rotation angle (up to 45 degrees), channel shift (changes gray scale intensity), width shift, height shift and zoom range. All five of these transformations were selected as they are all credible in the context of signature augmentation. Combining all these transformations, we expanded the dataset with 16 augmented images for each original ICDAR image, yielding a final dataset of 23,454 genuine images and 35,113 forgeries. This successfully improved the loss scores which in turn generated more convincing signatures when sampling from the latent dimension. This also allows for VAEs to be trained to learn a latent space representations for the signatures (genuine and forged) of a single person, as shown in Figure \ref{aug-training}.

\begin{figure}[!h]
    \begin{center}
        \includegraphics[width=0.36\linewidth]{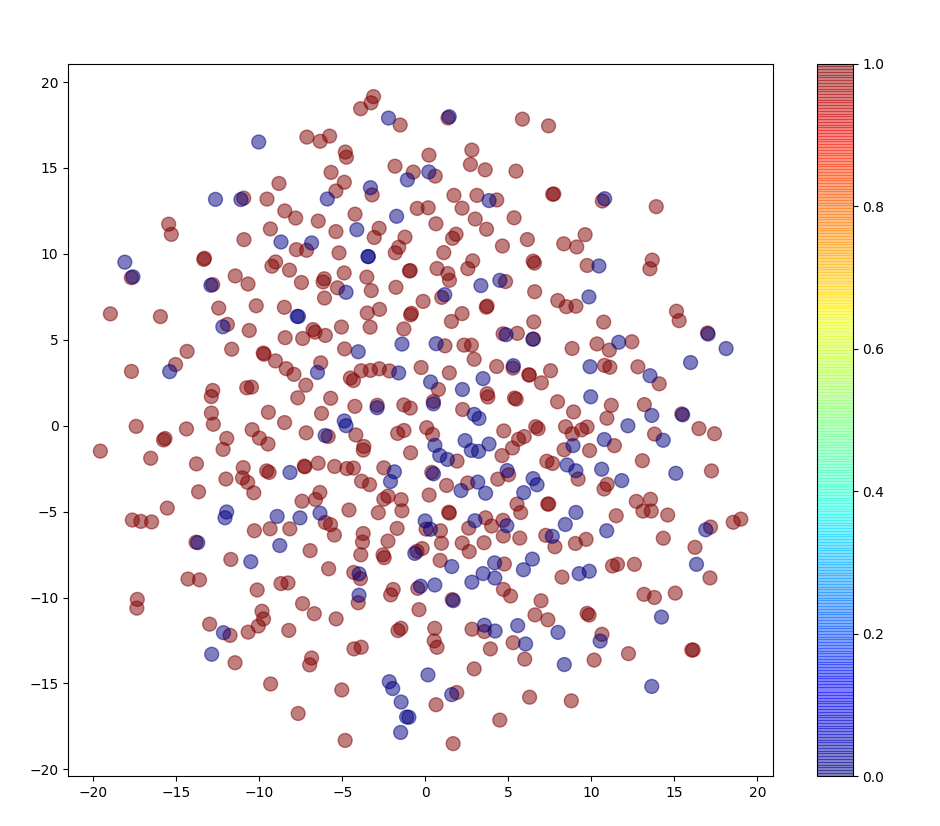}        \includegraphics[width=0.42\linewidth]{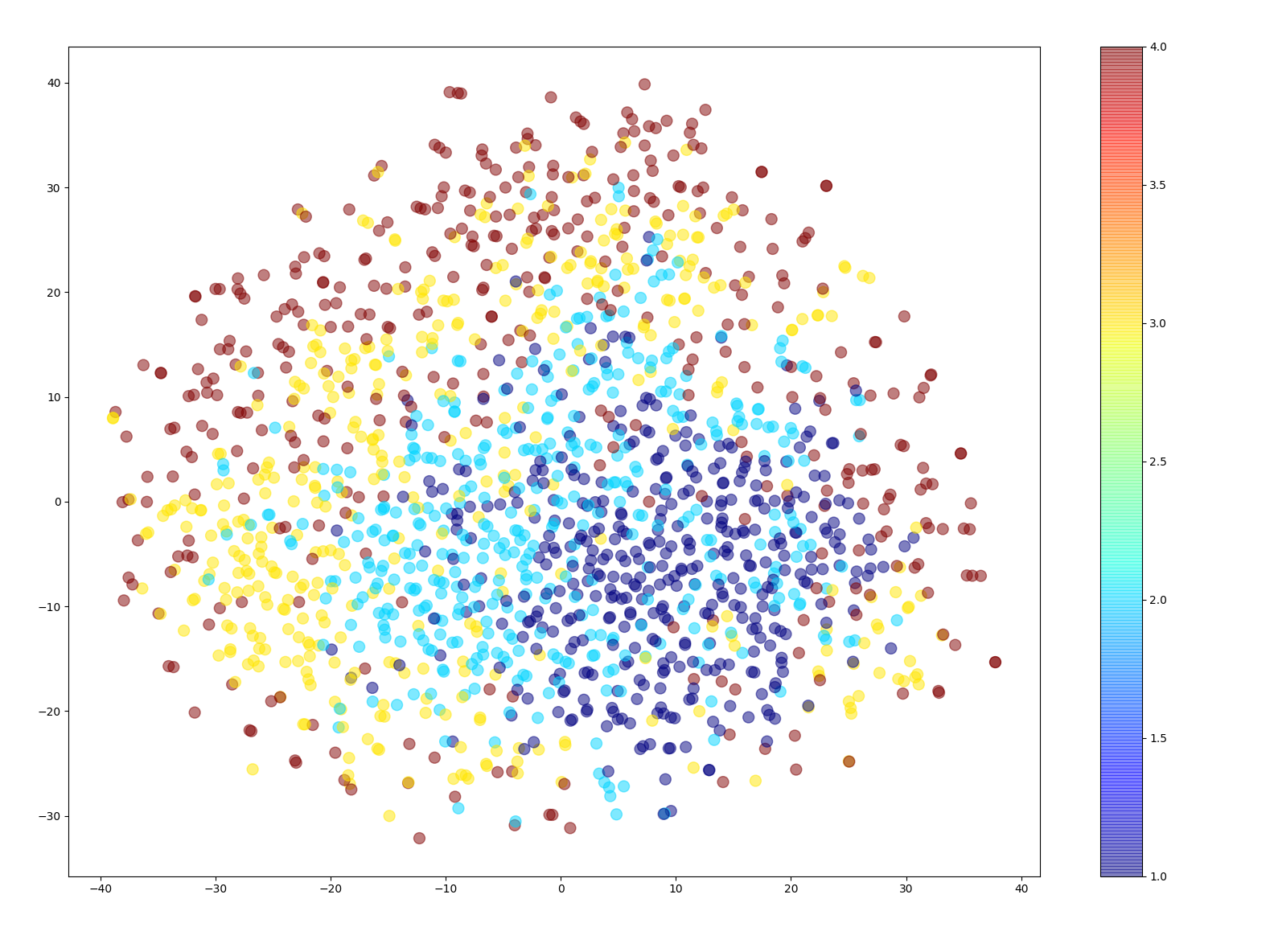} 
    	\caption{t-SNE representation of the 64-D latent vector of a real signature and its forgery (left) and one signature vs. three other people's signature (right)}
    	\label{tsne-plots}
    \end{center}
\end{figure}

From these VAE models we can now train kNN and Random Forest classifiers to detect forgeries from the latent space representations and from the reconstruction loss. The idea behind this is that we train the VAEs with genuine signatures only, such that the VAEs fail to reconstruct forgeries but successfully reconstruct genuine signatures. In order to visualise the differences between the real and fake latent space representations we can use t-SNE dimensionality reduction to generate the following plots detailed in Figure \ref{tsne-plots}.

Figure \ref{tsne-plots} shows how a real and forged latent representation looks like (mapped down to 2D) as seen on the left hand side of the image. Indeed, this looks challenging for a classifier in 2D but this may be a much easier task in the original 64-D of the latent space. By comparison, the right hand side of Figure \ref{tsne-plots} shows how it seems much easier to distinguish four different types of signatures from each other (i.e. identifying random forgeries from originals). 

\begin{figure}[!h]
    \begin{center}
        \includegraphics[width=0.76\linewidth]{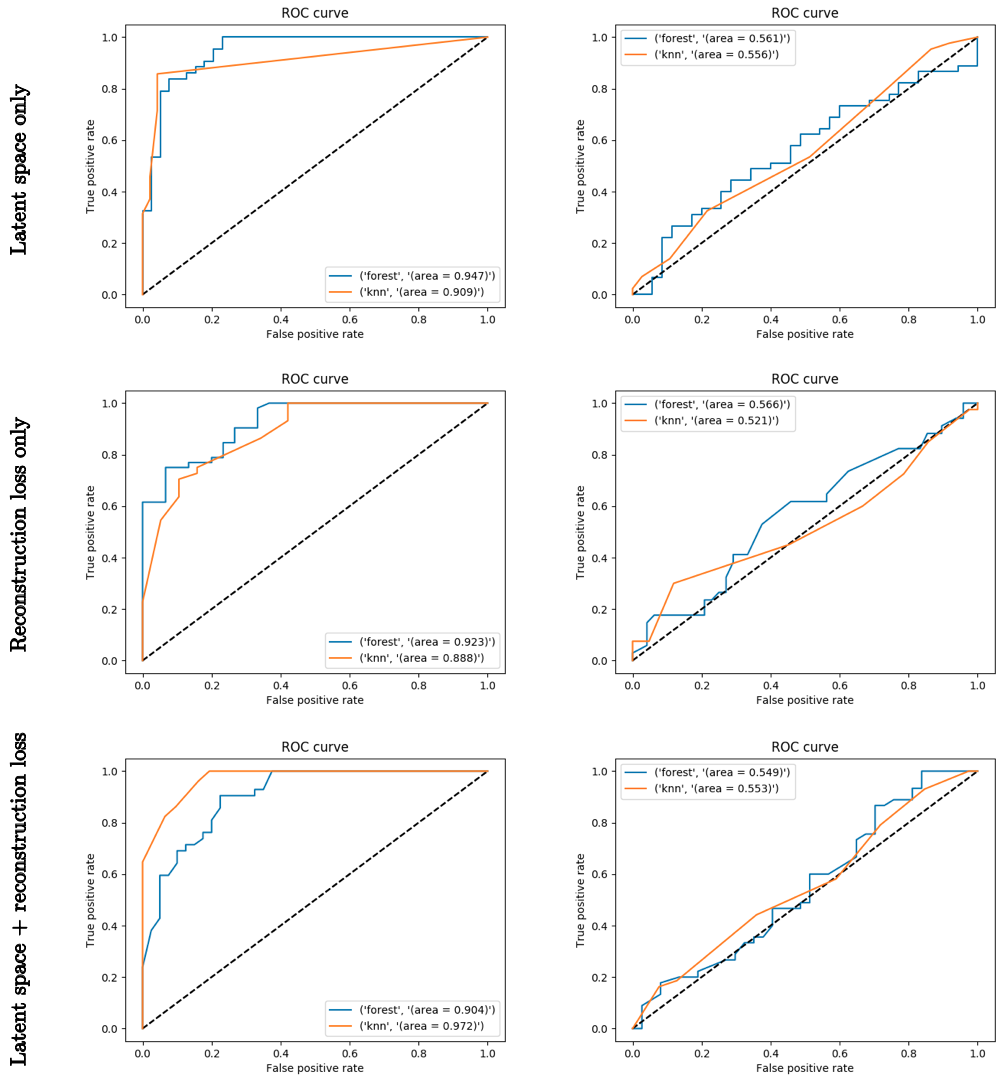}        
    	\caption{Performance ROC plots for easy (left) and hard signatures (right) using different features}
    	\label{ROC-plots}
    \end{center}
\end{figure}

The genuine/forgery classification performance of the kNN and Random Forest models can be seen in the following ROC plots as shown in Figure \ref{ROC-plots}. Figure \ref{ROC-plots} reveals that certain complex signatures are much harder to convincingly forge thus are easy to classify (left side). Similarly other signatures are much easier to forge and thus much more difficult to classify (right side). This effect is true no matter what we use as the classification input: latent space, reconstruction loss or even both.  

In addition, the summary of performances (accuracy, recall and F1) over the entire dataset is shown in Table 1 as follows. Table 1 shows that it seems the latent space information is equally as informative as the reconstruction loss. In addition, considering both simultaneously has no significant performance impact on genuine/forgery classification. Overall, the performances are nevertheless notably worse than existing alternatives. 

\begin{center}
 \begin{tabular}{||c c c c||} 
 \hline
   & Latent Space & Reconstruction Loss & Latent Space \& Reconstruction Loss \\ [0.5ex] 
 \hline\hline
 Accuracy & 0.6308 & 0.6237 & 0.6202 \\ 
 \hline
 Recall & 0.5494 & 0.6074 & 0.6040  \\
 \hline
 F1 score & 0.5671 & 0.6000 & 0.6011 \\ 
 \hline
\end{tabular}
\end{center}
\begin{center}
    Table 1: Classification performance over 63 signatures using latent space and reconstruction loss
\end{center}

A brief investigation was done to explore the potential of using MAML to improve weight initialisation. The goal of this was to reduce the number of signatures needed to train an accurate VAE for classification. However, this attempt of few-shot-learning was quickly observed to be flawed as the limited training set would result in a meta-learner that has overfit the original signatures. Thus the learned weight initialisations would not generalise to later signatures. As such, the investigation was instead shifted toward looking into VAE disentanglement and the possibility of posterior collapse.  

In an attempt to train a model that could learn a highly interpretable latent space, we investigated the possibility of integrating disentanglement to our VAE model. This consists in forcing the network to learn latent dimensions that are independent from each other by strictly restraining the network to the isotropic Normal distribution. That is, the second term of the loss function, the KL divergence loss, encourages the network to learn a latent space, described by the \texttt{mean} and the \texttt{std dev} vectors (see Figure \ref{system-diagram}), that is close to an isotropic Gaussian $\mathcal{N}(0,\mathbb{I})$. Therefore, because the covariance of an isotropic Gaussian is simply the identity matrix, this naturally encourages the different latent dimensions to be independent. However, the reconstruction loss might be too important and could prevent the model from efficiently learning a latent encoding featuring independent dimensions. One way to remedy this problem is to give more weight to the KL divergence loss through the introduction of a beta factor. Effectively, the loss from (5) becomes:
\begin{align*}
    \mathcal{L}(\phi; \mathbf{x}, \mathbf{z}) = \mathbb{E}_{q_{\phi} (\mathbf{z}|\mathbf{x})} [\log p (\mathbf{x}|\mathbf{z})] - \beta \cdot \mathbb{K} \mathbb{L}(q_{\phi} (\mathbf{z}|\mathbf{x}) || p(\mathbf{z}))
\end{align*}

When $\beta > 1$, the KL divergence loss takes more importance over the reconstruction loss. The beta term can then be adjusted just like any other hyperparameter. It effectively controls the trade-off between accurate reconstruction and high interpretability of the encoding. 

\begin{figure}[!h]
    \begin{center}
        \includegraphics[width=0.70\linewidth]{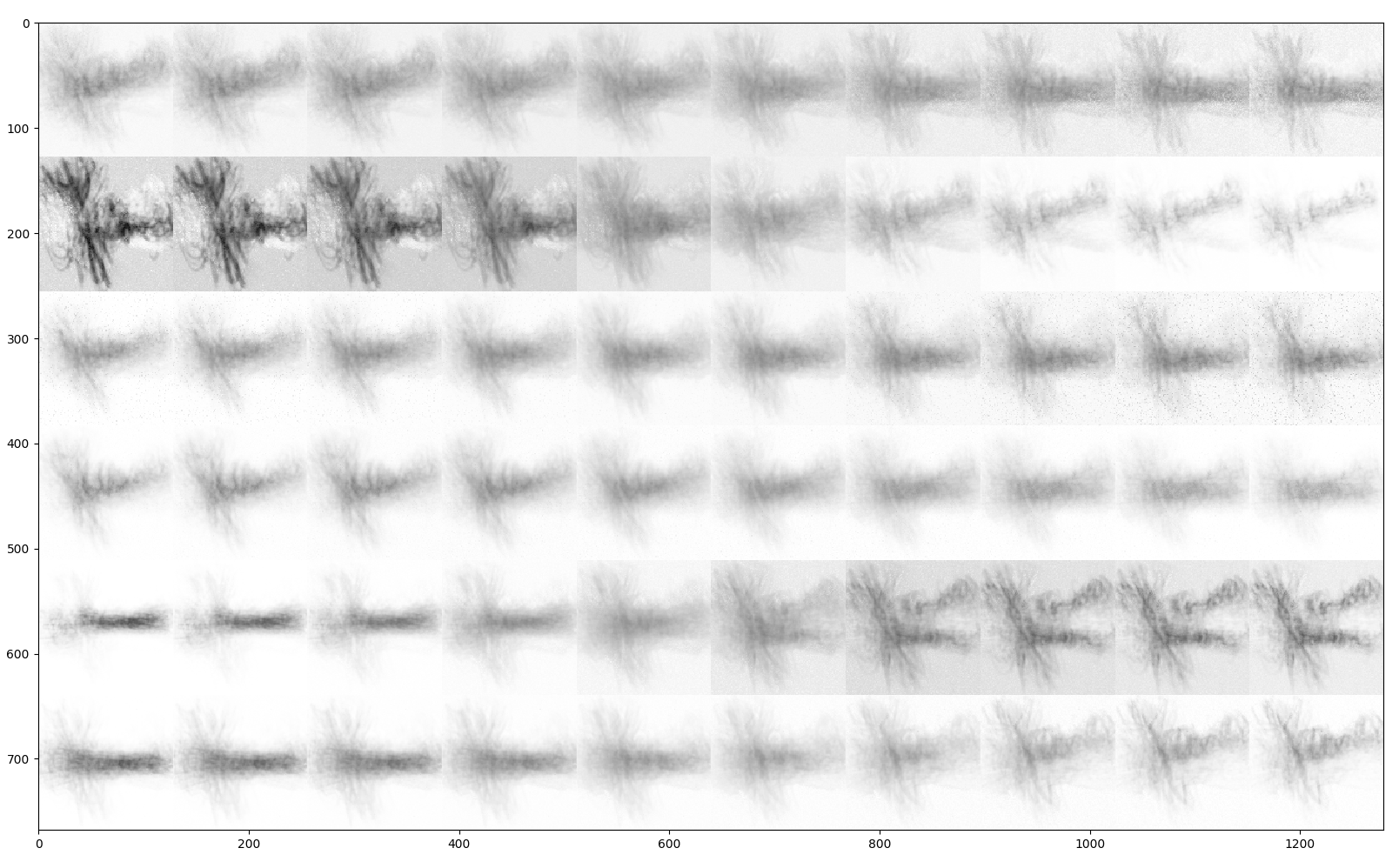}
    	\caption{Signatures generated from models with $\beta$ values 1, 1.25, 1.5, 1.75, 2 and 5 in rows 1 to 5 respectively. The sampling is done so that all dimensions are fixed to 0 except for the first one.}
    	\label{beta-vae-1}
    \end{center}
\end{figure}

\begin{figure}[!h]
    \begin{center}
        \includegraphics[width=0.70\linewidth]{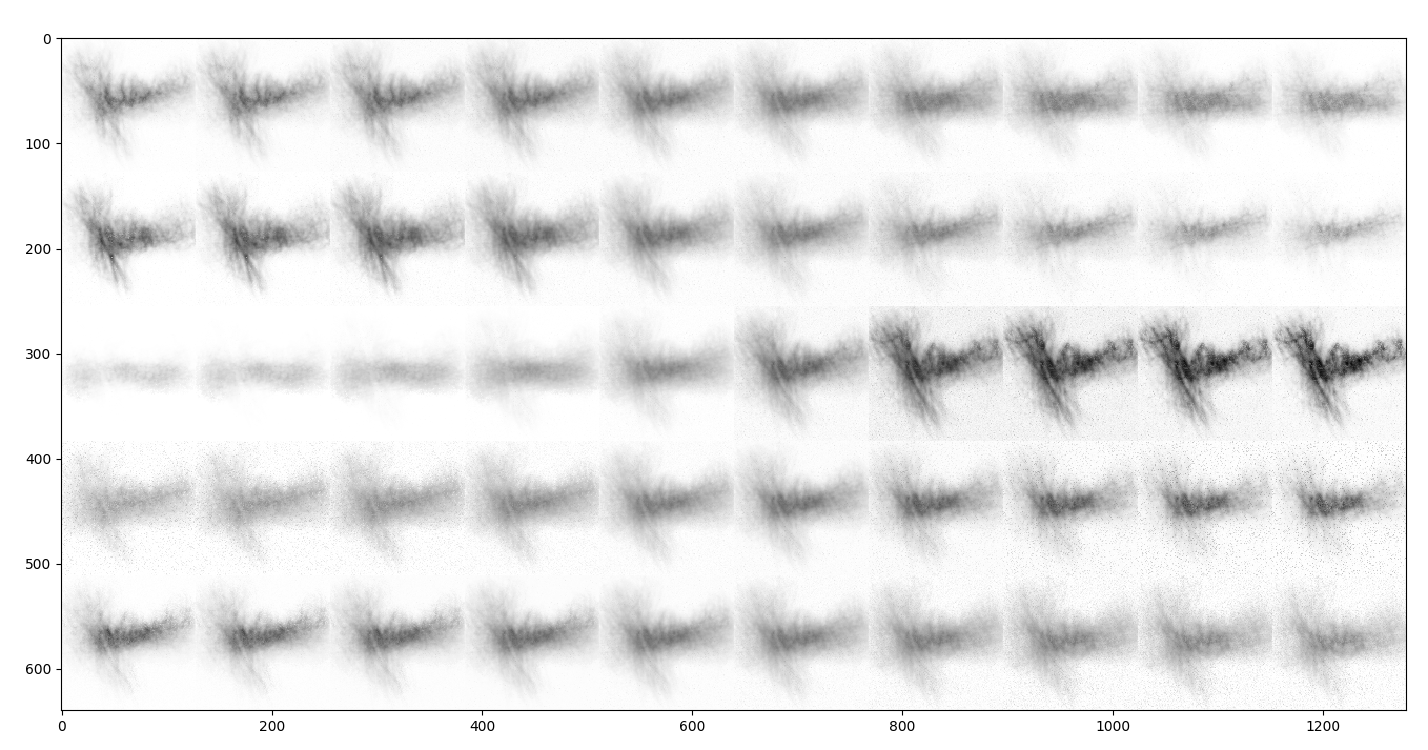}  
    	\caption{Samples from models with a $\beta$ of 1.75 and featuring a 5-D latent space. Each row represents variation on one dimension, 1 to 5 respectively, where all other dimensions are fixed to 0.}
    	\label{beta-vae-2}
    \end{center}
\end{figure}

\begin{figure}[!h]
    \begin{center}
        \includegraphics[width=0.70\linewidth]{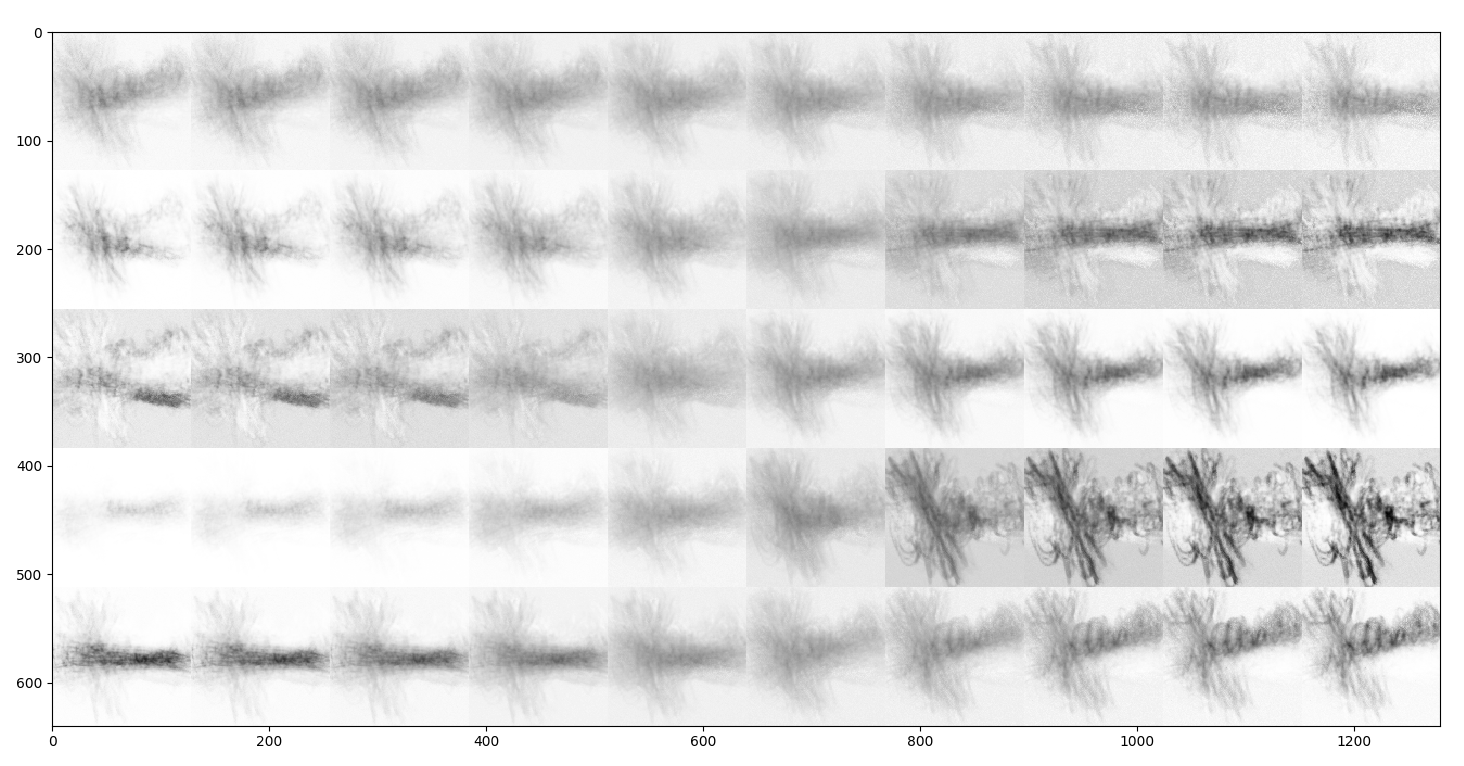}  
    	\caption{Samples from models with a $\beta$ of 1 (no disentanglement), featuring a 5-D latent space. Each row represents variation on one dimension, 1 to 5 respectively (all other dimensions fixed to 0).}
    	\label{beta-vae-3}
    \end{center}
\end{figure}

Six models with a 5-dimensional latent space and beta values of 1, 1.25, 1.5, 1.75, 2 and 5 were trained on signatures of 4 different participants. Figure \ref{beta-vae-1} shows images generated from those models where the sampling was done with all dimensions fixed to zero except for the first dimension which shifted between -4 and 4. The fourth row, sampled from the model with a beta value of 1.75, shows promising results since the latent variable seems to clearly encode the rotation of the signature. Figure \ref{beta-vae-2} shows how variation of each dimension of that model's latent space affects the generated signature. We see that each dimension seems to encode a specific attribute of the image which suggest that disentanglement did allow to decorrelate the latent dimensions. For example,  the third dimension seems to encode how much vertical space the signature takes and the fifth dimension seems to encode how predominant the center of the signature is compared to the edges. This is further supported by the fact that the different latent dimensions of the model trained with a beta value of 1 do not seem to clearly encode for one aspect of the image as seen on Figure \ref{beta-vae-3}. Indeed dimensions 2, 3 and 4 all seem to encode for the background color and how predominant the center of the signature is compared to the edges.  

\section{Investigating the possibility of posterior collapse} \label{section-posterior}

\subsection{Why care about posterior collapse?}
A common problem when training VAEs with high-capacity decoders is the problem of \textit{posterior collapse}, characterized by the network falling into the undesirable local optimum where $q_{\phi}(z|x) = p_{\theta}(z|x) = p(z)$, with a common assumption being $p(z) \sim N(0,I)$ \cite{he2019lagging}. This amounts to the latent code $z$ being "ignored". He, Spokoyny et al. in \cite{he2019lagging} hypothesize that this occurs because $z$ and $x$ will begin the training procedure as independent of each other, and that the KL term of the loss function may overwhelm the $\log(p_\theta(x))$ term at the start of training. Once the KL loss has become close to zero (necessitating that $q_{\phi}(z|x) \approx p_{\theta}(z|x) \approx p(z)$), the remainder of the training procedure serves to find model parameters which capture $p(x)$ without making use of $z$. This would evidently present a problem for classification on the latent vectors. Indeed, performance should be poor if little or no information is being captured in the latent space.

While this issue appears to be most commonly reported in powerful autoregressive decoders \cite{he2019lagging}, which are not being used in our architecture, we thought that the possibility of posterior collapse was worth investigating. It should inform whether our classifier results ought to be considered meaningful, or are more likely to reflect over-fitting to noise in a relatively information-poor latent space.

He, Spokoyny et al. in \cite{he2019lagging} prove the existence of posterior collapse by plotting the means of $q_{\phi}(z|x)$ and $p_{\theta}(z|x)$ against each other throughout training. They are able to do this on a synthetic dataset for which the expected value of $p_{\theta}(z|x)$ is approximated through discretization and MH sampling. Our investigation, while not as involved, examines the proportion of loss attributable to the KL term rather than the reconstruction loss throughout the course of training. This allows us to make a reasonable guess as to whether we should suspect posterior collapse. 

\subsection{Training loss breakdown and discussion}
The observed pattern of the average loss breakdown over the course of training does not match what we would necessarily expect from the He, Spokoyny et al.'s hypothesis, where we would expect to see a rapid decline of KL divergence, followed by a steady decrease in reconstruction loss.

\begin{figure}[!htb]
\begin{center}
    \includegraphics[width=0.75\linewidth]{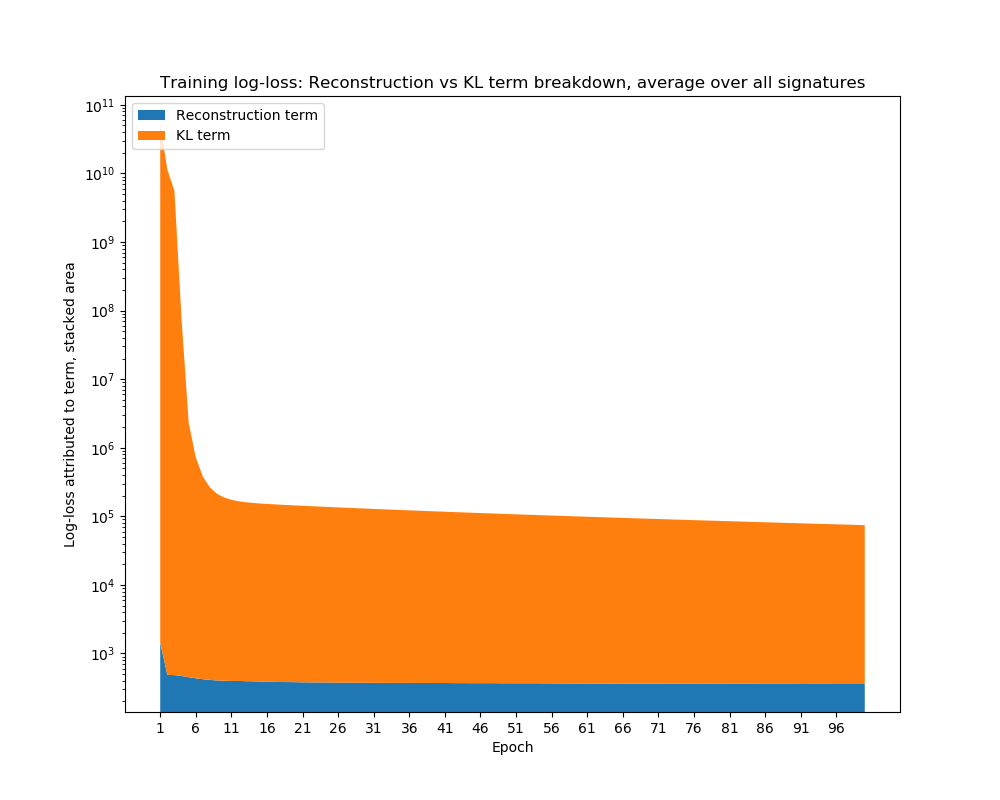}
    \caption{Breakdown between the reconstruction (MSE) and the KL components of the VAE loss.}
    \label{fig:avg_kl_mse}
\end{center}
\end{figure}

However, from Figure \ref{fig:avg_kl_mse} we note that the KL term does decline steeply in the first 10 or so epochs. Reconstruction loss then remains relatively constant over training while the KL term declines slowly, even while it continues to make up the majority of the total loss in some instances.

Examining Figure \ref{fig:indiv_charts}, the individual loss plots for specific signatures reveal that certain signatures maintain a large KL divergence (i.e. signatures 3, 6, 13), while others have a KL divergence that seems to taper off asymptotically to 0 (signatures 1, 9, and 12). A KL divergence of zero indicates that the learned $q(z|x) \approx p(z)$, leading us to suspect posterior collapse in some signatures.

A potential strategy for addressing posterior collapse could be to try placing a small-valued $\beta$ coefficients (i.e. 0.1, 0.01, etc) on the KL divergence term, to prevent the term from overwhelming the loss. We re-ran our experiments after training our VAEs with $\beta = 0.001$ and found marginally better performance than the results (with default $\beta = 1.0$) reported in Section 4, but further validation would be required to ensure that this didn't arise due to random chance.

\begin{center}
 \begin{tabular}{||c c c||} 
 \hline
   & Latent Space, $\beta=1.0$ & Latent Space, $\beta=0.001$ \\ [0.5ex] 
 \hline\hline
 Accuracy & 0.6308 & 0.6907 \\ 
 \hline
 Recall & 0.5494 & 0.6429  \\
 \hline
 F1 score & 0.5671 & 0.6647 \\ 
 \hline
\end{tabular}
\end{center}
\begin{center}
    Table 2: Average classification performance over all signatures, comparing $\beta = 1.0$ vs $\beta = 0.001$
\end{center}

The performance is still relatively poor, indicating that a constant weight on the KL regularization term may not be the correct solution. It would worth investigating the KL-term cost annealing approach used by Bowman et al. (2015) \cite{bowman2015generating}, where the KL weight increases over time.
\newcommand\littlefigscale{0.12}
\begin{figure}[!htb]
\begin{center}
    \includegraphics[scale=\littlefigscale]{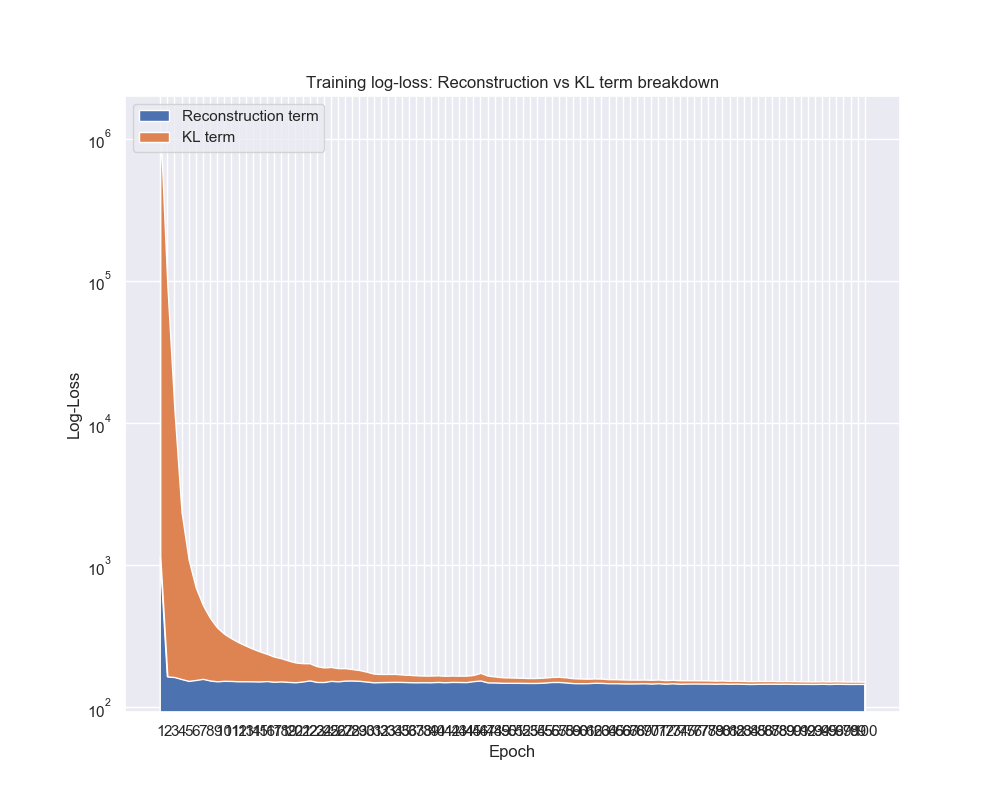}
    \includegraphics[scale=\littlefigscale]{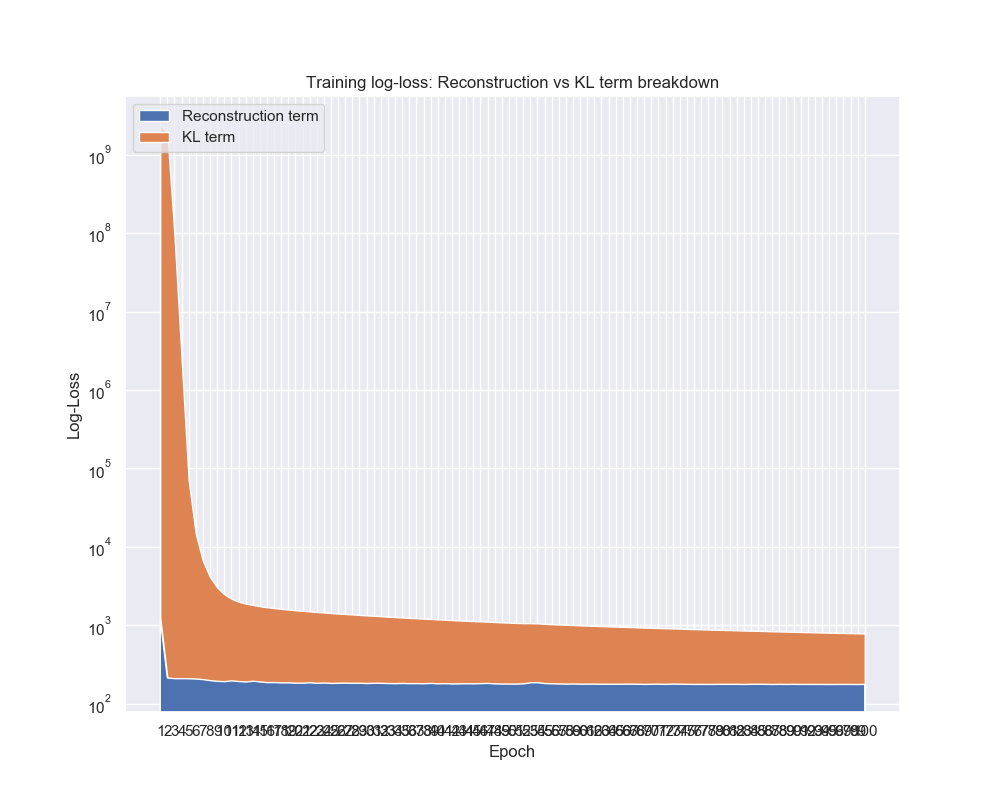}
    \includegraphics[scale=\littlefigscale]{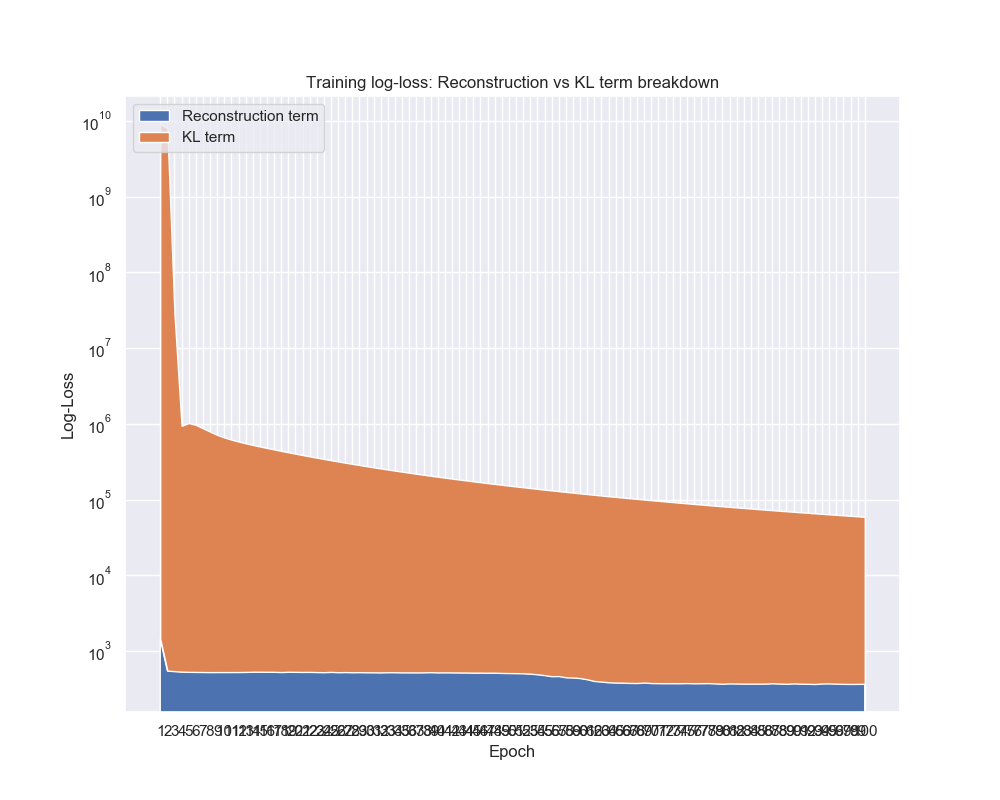}
    \includegraphics[scale=\littlefigscale]{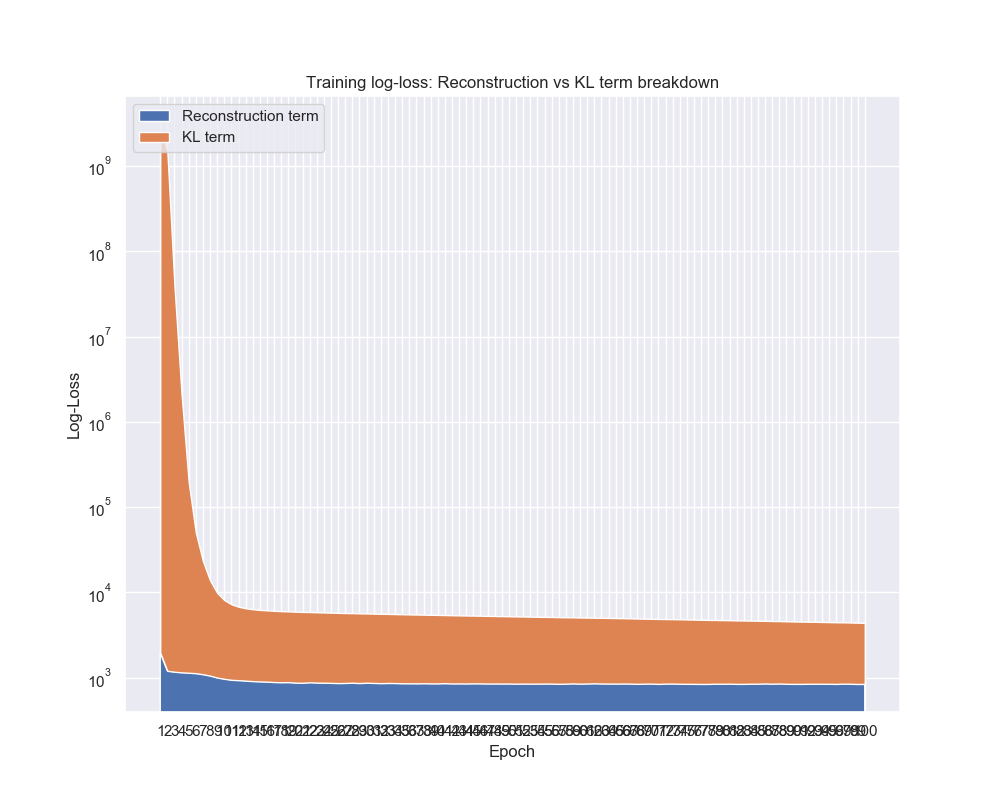}
    \\
    \includegraphics[scale=\littlefigscale]{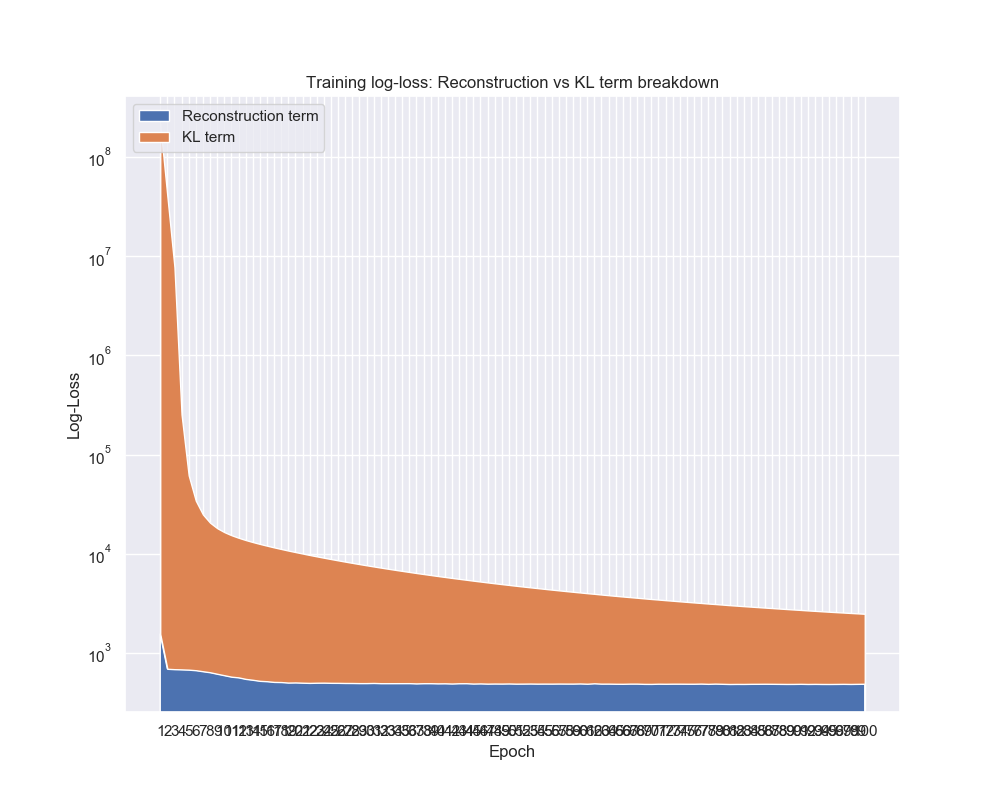}
    \includegraphics[scale=\littlefigscale]{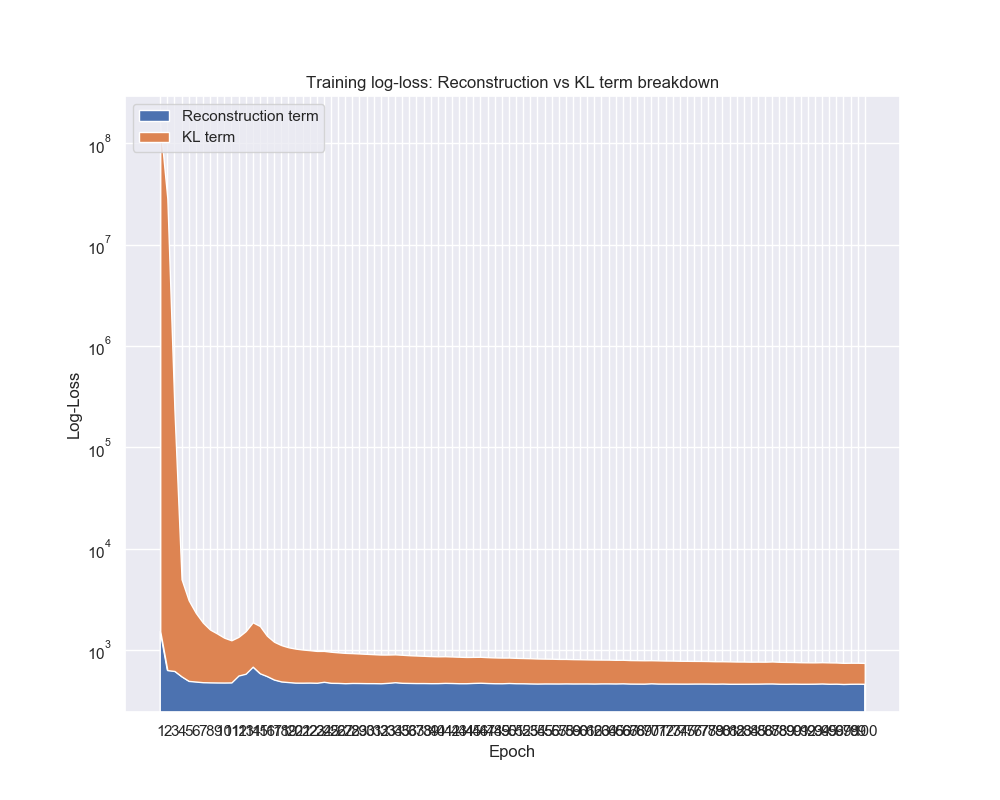}
    \includegraphics[scale=\littlefigscale]{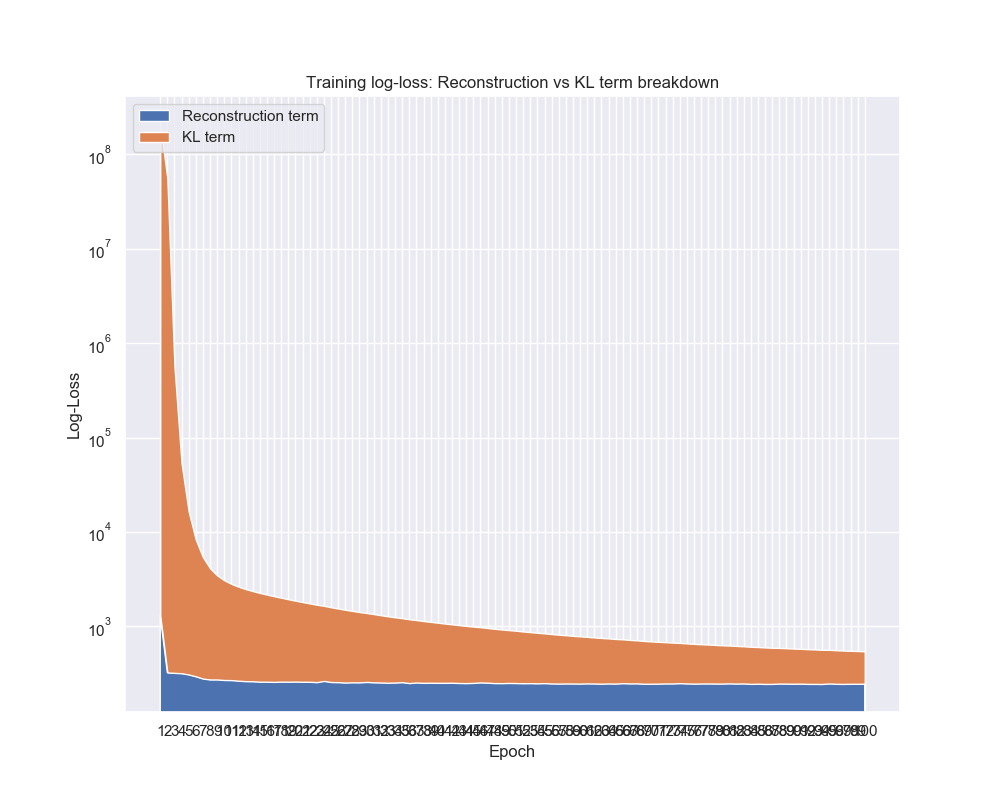}
    \includegraphics[scale=\littlefigscale]{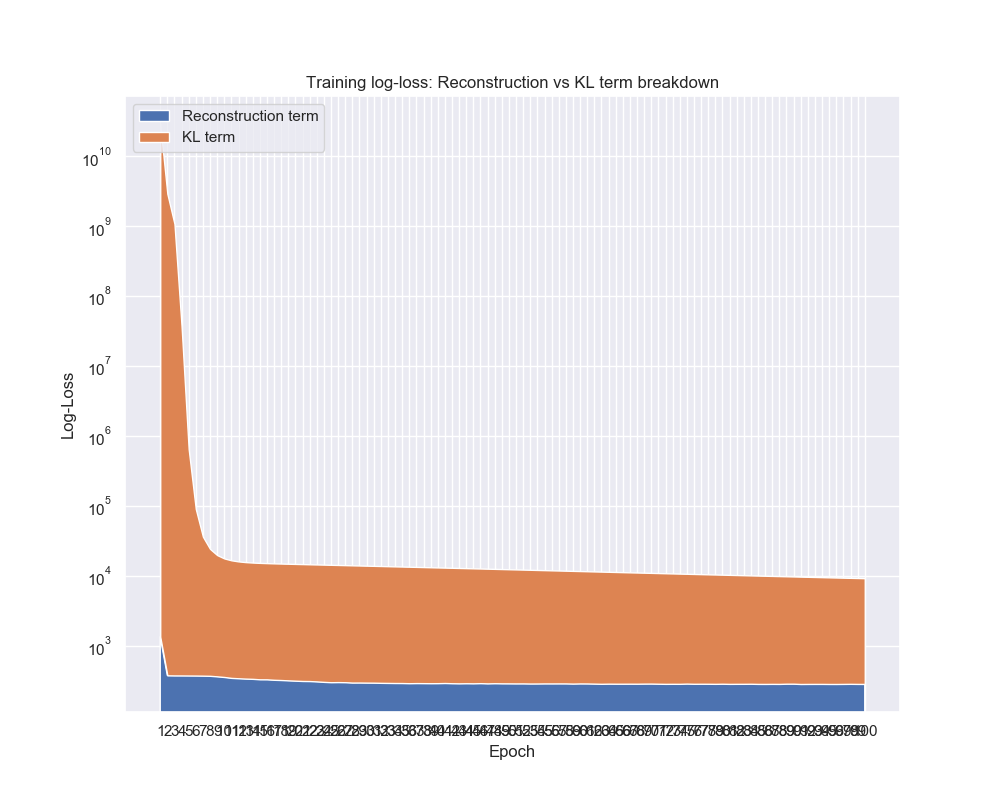}
    \caption{Examples of individual loss graphs for (in order) signatures 1, 2, 3, 4, 6, 9, 12, and 13 (indices are as provided by the dataset). Orange region = KL term, blue = reconstruction term.}
    \label{fig:indiv_charts}
\end{center}
\end{figure}

\section{Limitations and Future Work}

Various limitations were encountered throughout this project. The most notable limitation is that of the amount of data required for a generative model, such as the VAE used, to generate crisp and representative samples from the latent space. As shown in section 4, this results in samples that simply look like a normal distribution of inputs. However, using some data augmentation techniques this can be somewhat addressed but not fully. 

Another limitation of the results presented here is the practical application limitation. Indeed, offline signature forgery by hand has become somewhat more rare in recent years. Modern forgers attempting skilled forgery offline predominantly use digital cameras, photo editing and electronic means to replicate signatures. Because of this, the most common issue is no longer that the offline signature does not match, but rather that it was copy-pasted by an author other than the original signature holder. As such, the implications of the work presented here is perhaps more relevant for online signature verification where the timing and intensity of the signature over time is measured during the signing process. 

Future work could look at leveraging the techniques outlined in section 4 to complement existing online signature verification ensembles. It seems that the shortcomings of this method may be drastically different from those of other methods that look at direct classification without considering the latent space and reconstruction loss. 

In addition, another possibility for future work could be to help further address the small-data problem encountered here. Indeed, the key advantage of the VAE architecture detailed in this paper is that it may also be modified and used to generate synthetic signatures for additional data augmentation. 

\section{Conclusions}

This paper's main conclusions centre around the demonstration of how a VAE can be modified to handle the ICDAR dataset and generate latent representations of signatures using augmented data for one-class classification. All the code developed during this project has been open-sourced and is publicly available on GitHub \cite{viscardi2019catch}. While the performances reported in this paper are not anywhere near exceeding that of the current ICDAR competition winners for offline verification, this paper does nonetheless highlight the possibilities and potential of VAEs for one-class classification. 

In addition, this paper shows why a VAE approach achieves performances of no better than chance on certain difficult signatures. This is followed up with further experimental evidence looking at how posterior collapse arises during this VAE-based approach, and suggests that a simple constant weighting on the KL regularization term is insufficient for achieving a meaningful latent code. 

Indeed after a brief review, for offline signature datasets like ICDAR, few-shot-learning and meta-learning methods like MAML seem unfeasible due to the small data problem.


\bibliographystyle{plain}
\bibliography{main}

\end{document}